\documentclass[sigconf]{acmart}

\usepackage{multirow}
\usepackage{bm}
\usepackage{balance}

\AtBeginDocument{%
  }

\copyrightyear{2023}
\acmYear{2023}
\setcopyright{acmlicensed}\acmConference[MM '23]{Proceedings of the 31st ACM International Conference on Multimedia}{October 29-November 3, 2023}{Ottawa, ON, Canada}
\acmBooktitle{Proceedings of the 31st ACM International Conference on Multimedia (MM '23), October 29-November 3, 2023, Ottawa, ON, Canada}
\acmPrice{15.00}
\acmDOI{10.1145/3581783.3612363}
\acmISBN{979-8-4007-0108-5/23/10}

\begin{document}

\title{3DStyle-Diffusion: Pursuing Fine-grained Text-driven 3D Stylization with 2D Diffusion Models}

\author{Haibo Yang}
\authornote{Haibo Yang and Yang Chen contributed equally to this work.}
\affiliation{%
  \institution{School of Computer Science, Fudan University}
  	    \country{China}
}
\email{yanghaibo.fdu@gmail.com}

\author{Yang Chen}
\authornotemark[1]
\affiliation{%
  \institution{University of Science and Technology of China}
      \country{China}
}
\email{c1enyang.ustc@gmail.com}

\author{Yingwei Pan}
\affiliation{%
	\institution{University of Science and Technology of China}
	    \country{China}
}
\email{panyw.ustc@gmail.com}

\author{Ting Yao}
\affiliation{%
	\institution{HiDream.ai Inc.}
	    \country{China}
}
 \email{tingyao.ustc@gmail.com}

\author{Zhineng Chen}
\authornote{Corresponding Author.}
\affiliation{%
	\institution{School of Computer Science, Fudan University}
	    \country{China}
}
\email{zhinchen@fudan.edu.cn}

\author{Tao Mei}
\affiliation{%
	\institution{HiDream.ai Inc.}
	    \country{China}
}
\email{tmei@hidream.ai}

\renewcommand{\shortauthors}{Haibo Yang et al.}

\begin{abstract}
  3D content creation via text-driven stylization has played a fundamental challenge to multimedia and graphics community. Recent advances of cross-modal foundation models (e.g., CLIP) have made this problem feasible. Those approaches commonly leverage CLIP to align the holistic semantics of stylized mesh with the given text prompt. Nevertheless, it is not trivial to enable more controllable stylization of fine-grained details in 3D meshes solely based on such semantic-level cross-modal supervision. In this work, we propose a new 3DStyle-Diffusion model that triggers fine-grained stylization of 3D meshes with additional controllable appearance and geometric guidance from 2D Diffusion models. Technically, 3DStyle-Diffusion first parameterizes the texture of 3D mesh into reflectance properties and scene lighting using implicit MLP networks. Meanwhile, an accurate depth map of each sampled view is achieved conditioned on 3D mesh. Then, 3DStyle-Diffusion leverages a pre-trained controllable 2D Diffusion model to guide the learning of rendered images, encouraging the synthesized image of each view semantically aligned with text prompt and geometrically consistent with depth map. This way elegantly integrates both image rendering via implicit MLP networks and diffusion process of image synthesis in an end-to-end fashion, enabling a high-quality fine-grained stylization of 3D meshes. We also build a new dataset derived from Objaverse and the evaluation protocol for this task. Through both qualitative and quantitative experiments, we validate the capability of our 3DStyle-Diffusion. Source code and data are available at \url{https://github.com/yanghb22-fdu/3DStyle-Diffusion-Official}.

\end{abstract}

\ccsdesc[500]{Information systems~Multimedia content creation}

\keywords{Text-driven 3D Stylization, Diffusion Model, Depth}

\begin{teaserfigure}
\vspace{-0.2in}
\centering
  \includegraphics[width=0.93\textwidth]{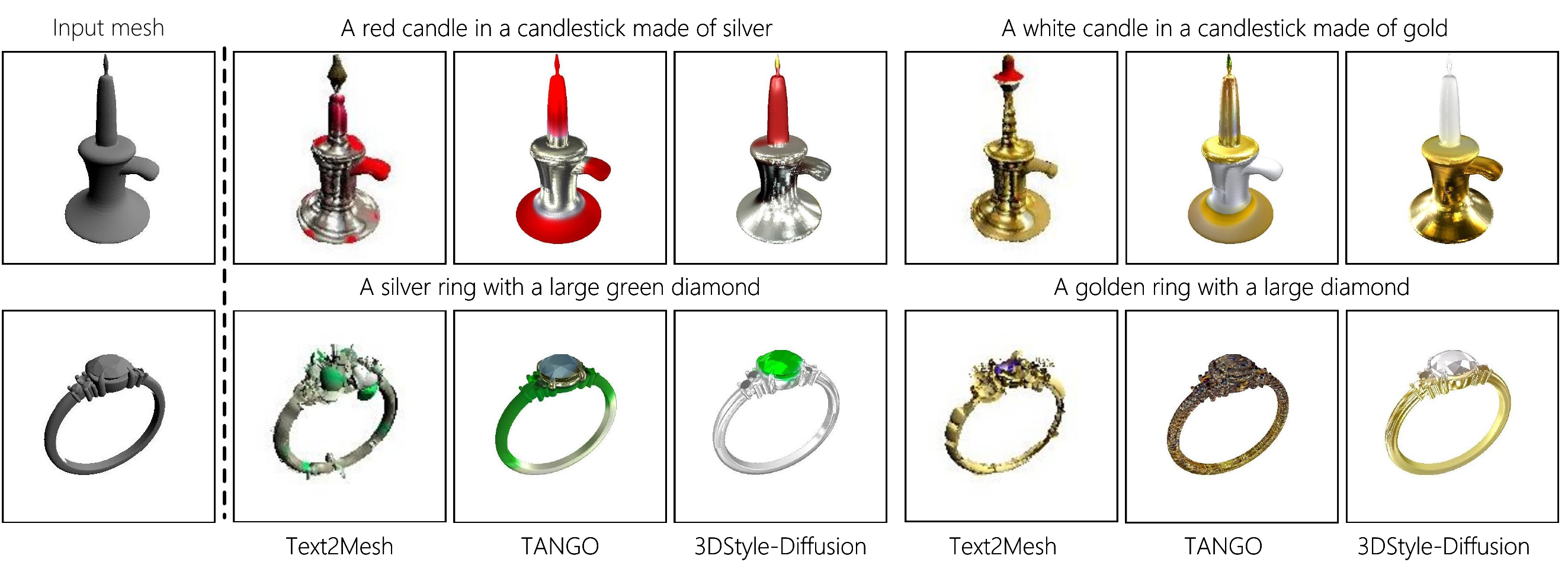}
  \vspace{-0.10in}
  \caption{Existing CLIP-based text-driven 3D stylization techniques (e.g., Text2Mesh \cite{Michel_2022_CVPR-Text2Mesh} and TANGO \cite{chen2022tango}) often fail to enable a precise stylization of fine-grained details. Our 3DStyle-Diffusion instead exploits controllable appearance and geometric guidance from 2D Diffusion models, leading to a more fine-grained text-driven stylization of 3D meshes.}
  \label{fig:teaser}
\end{teaserfigure}


\maketitle

\section{Introduction}
3D content creation via stylization is one of the fundamental directions in multimedia and graphics community, which targets for stylizing primary 3D meshes according to the given text prompts \cite{Michel_2022_CVPR-Text2Mesh,chen2022tango} or visual prompts (images \cite{pavllo2020convmesh, kato2018neural}  or shapes \cite{hanocka2018alignet,liu2018paparazzi}). This direction triggers automatic style-specific editability of 3D assets, thereby playing a critical role in numerous applications. In between, text-driven 3D stylization might be the most challenging. The complex and fuzzy gap between visual content and text prompt makes it extremely hard to precisely stylize the visual appearance and geometry of 3D meshes solely conditioned on text prompt.

Taking the inspiration from recent advance of cross-modal foundation models that facilitate a series of vision-language tasks \cite{pan2022auto,li2021scheduled,weng2023transforming,wang2022omnivl}, recent pioneering practices start to leverage pre-trained Contrastive Language-Image Pre-training (CLIP) model \cite{radford2021clip} to bridge the cross-modal gap in text-driven 3D stylization. In particular, Text2Mesh \cite{Michel_2022_CVPR-Text2Mesh} learns to predict stylized color and geometric information of each mesh vertex based on input text prompt by using CLIP-based holistic semantic similarity as supervision. TANGO \cite{chen2022tango} further upgrades the modeling of texture on 3D meshes by exploiting more factors (e.g., lighting conditions and reflectance properties). But it still leverages CLIP-based holistic semantic loss between rendered images and input text prompt. Although promising results are achieved for text-driven 3D stylization of meshes in arbitrary topology, these techniques with CLIP-based supervision mostly fail to enable a precise controllable stylization of fine-grained details in 3D meshes. As shown in Figure \ref{fig:teaser} (the last example with text prompt ``a golden ring with a large diamond''), TANGO basically produces a fully golden ring, but mistakenly stylizes the fine-grained components (e.g., the diamond).

In this work, we propose to mitigate this issue from the viewpoint of exploiting 2D Diffusion models that provide more controllable appearance and geometric supervisory signals to guide the learning, pursuing more fine-grained text-driven stylization of 3D meshes. The CLIP-based supervision holistically aligns visual appearances of rendered images and text prompts. In contrast, our pre-trained controllable 2D Diffusion model (ControlNet \cite{zhang2023adding}) further enhances the learning of rendered images with more conditions in a diffusion process, including both text prompt and precise geometric information (depth maps) derived from primary 3D mesh. This way simultaneously manifests the emphasis of both holistic semantic and local geometric consistency with regard to text prompt and primary 3D mesh, thereby yielding high-quality 3D stylization especially over fine-grained details. For example, in Figure \ref{fig:teaser}, our work nicely performs fine-grained 3D stylization of each component.

By consolidating the idea of integrating 3D stylization with controllable 2D Diffusion models, we design a novel diffusion model, namely 3DStyle-Diffusion, for text-driven 3D stylization. Our launching point is to unify the rendering of stylized 3D meshes and 2D diffusion process of controllable image synthesis into an end-to-end scheme. Technically, 3DStyle-Diffusion first capitalizes on implicit MLP networks to parameterize the texture of 3D mesh into scene lighting and reflectance properties, leading to rendered images through ray casting \cite{roth1982ray}. During the ray casting process, we also achieve a precise depth map of each sampled view derived from primary 3D mesh. After that, the rendered image is fed into a pre-trained 2D diffusion model (Control Stable Diffusion) to trigger controllable image generation conditioned on both text prompt and depth map, leading to fine-grained text-driven 3D stylization.

In summary, we have made the following contributions:
(\textbf{I}) 3DStyle-Diffusion enables fine-grained text-driven 3D stylization by guiding image rendering with controllable 2D diffusion model in an end-to-end manner.
(\textbf{II}) To evaluate the challenging text-driven 3D stylization over fine-grained details, we build a new dataset (namely Objaverse-3DStyle) derived from Objaverse \cite{deitke2022objaverse} that contains various 3D composite assets (with at least 2 different components). A new evaluation protocol is also designed to quantitatively evaluate all methods.
(\textbf{III}) Both qualitative and quantitative experiments are performed over our newly collected dataset, which demonstrate the effectiveness of 3DStyle-Diffusion.

\begin{figure*}[th!]
\begin{center}
\includegraphics[width=1.0\linewidth]{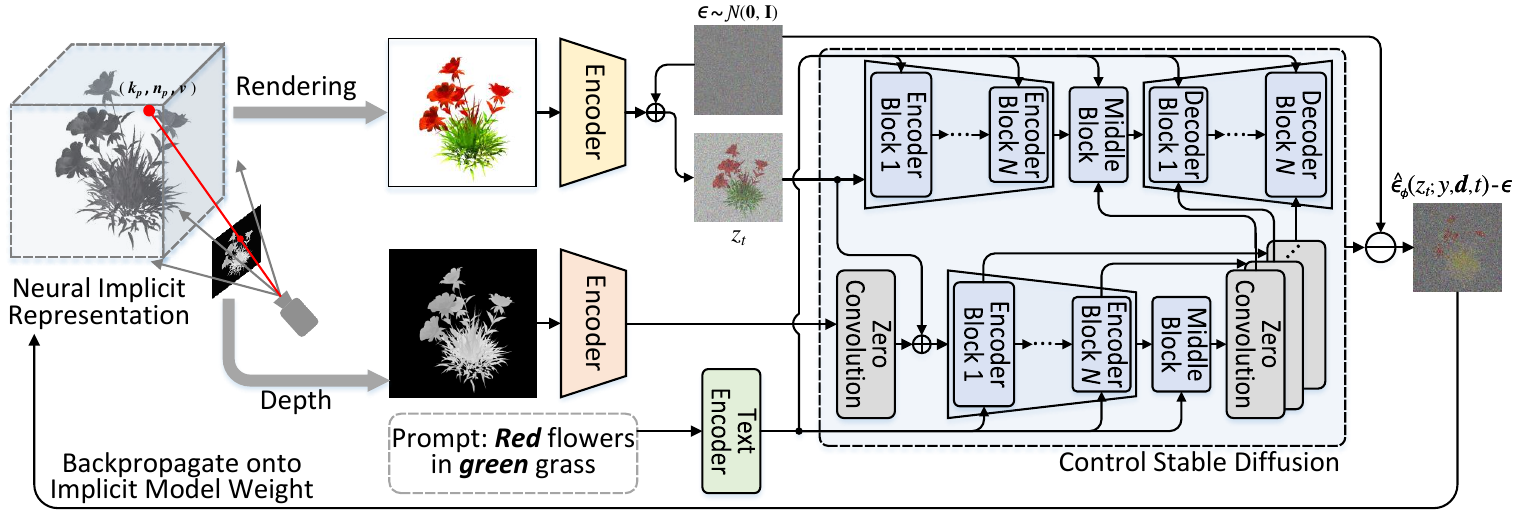}
\end{center}
\vspace{-0.15in}
  \caption{An overview of 3DStyle-Diffusion. 3DStyle-Diffusion nicely triggers fine-grained stylization of 3D meshes based on text prompt (e.g., ``Red flowers in green grass''). Specifically, 3DStyle-Diffusion utilizes MLP networks to parameterize the texture of the mesh 
  into reflectance properties and scene lighting. Given a camera position \bm{$c$}, rays cast from the camera to the mesh surface intersection points \bm${k_p}$. Next, \bm${k_p}$ and its norm \bm${n_p}$ plus the ray direction \bm{$v$} are fed into the implicit representation networks. A fully differentiable renderer is used to obtain 
  the rendering image \bm{$x$}. Meanwhile, we achieve the depth map of 3D mesh in the given view during the ray casting process. Then, the rendering image \bm{$x$} is diffused and reconstructed using a pre-trained control
  stable diffusion model conditional on the depth map, the time step, and the input text prompt to predict the injected noise, which is then used to backpropagate gradients to update the parameters of the implicit MLP network. This way unifies both image rendering and diffusion processes into an end-to-end framework, leading to fine-grained 3D stylization.}
\label{fig:framework}
\vspace{-0.2in}
\end{figure*}

\section{RELATED WORK}
\textbf{Text-to-image (T2I) Synthesis.}
Earlier research in T2I synthesis was dominated by GANs \cite{goodfellow2020generative, mirza2014conditional, reed2016generative, esser2020taming}, which employed a two-player minimax game mechanism to train the generator to produce synthetic images indistinguishable from real images. Recently, Diffusion models \cite{sohl2015deep, ho2020ddpm, nichol2021improved, ho2022classifier, rombach2022stablediffusion} have emerged as the new trend of generative models for generating high-quality images. With the help of large-scale image-text paired datasets, diffusion models have been widely exploited to form state-of-the-art T2I models (such as DALL-E2 \cite{ramesh2022hierarchical} and Imagen \cite{saharia2022photorealistic}). These models can generate images that are closely aligned with the input text prompt. 
Motivated by these successes, many works attempt to utilize pre-trained T2I diffusion models for various tasks such as text-driven image editing \cite{brooks2022instructpix2pix, kawar2022imagic, mokady2022null, ruiz2022dreambooth, nichol2021glide, huang2022draw}. However, using 2D diffusion model to achieve fine-grained text-driven 3D stylization is seldom explored and remains an open problem in the multimedia and vision field.

\textbf{Text-driven 3D Generation and Manipulation.}
Recently, significant advancements have been made in multimedia content creation \cite{ho2020ddpm, ramesh2022hierarchical, saharia2022photorealistic, chen2019mocycle, chen2019animating, poole2022dreamfusion, jain2022zero, pan2017create, zhang2023}. With notable advancements in text-image cross-modal models, there has been a growing interest in text-driven 3D visual synthesis. These works can be generally categorized into two groups. The first group is text-to-3D generation. Pioneering works DreamField \cite{jain2022zero} and CLIP-Mesh \cite{mohammad2022clip} leverage cross-modal knowledge from a pre-trained image-text model (i.e., CLIP \cite{radford2021clip}) to optimize the underlying 3D representations (NeRFs and Meshes). By doing so, they eliminate the need for training data, thus enhancing the efficiency and convenience of 3D content creation. Sparked by the success of diffusion models in T2I generation, recent works  \cite{poole2022dreamfusion, wang2022score, lin2022magic3d, metzer2022latent, nichol2022pointe} utilize pre-trained T2I diffusion models for text-to-3D generation, yielding impressive~results.

Another group is mesh-based text-driven 3D stylization, which has also drawn increasing research attention due to its wide applicability. Recent works have made impressive progress in automating the process of 3D stylization only using text prompts. A pioneering practice in text-driven 3D stylization is Text2Mesh \cite{Michel_2022_CVPR-Text2Mesh}. Given an input mesh and a text prompt, Text2Mesh predicts stylized color and displacement for each mesh vertex by leveraging an off-the-shelf, pre-trained CLIP model. Then the stylized mesh is directly achieved through this colored vertex displacement procedure. TANGO \cite{chen2022tango} further incorporates reflectance properties and scene lighting to improve the realism of text-driven 3D mesh stylization.

Our work falls into the latter group of text-driven 3D stylization. Though existing works (Text2Mesh and TANGO) are able to generate promising text-driven stylized meshes through CLIP-based supervision, they still fail to enable a precise controllable stylization of fine-grained details in 3D meshes. In contrast, our work novelly facilitates 3D stylization with controllable 2D Diffusion models, pursuing more fine-grained text-driven stylization of 3D meshes.

\section{METHOD}

In this section, we elaborate our proposed 3DStyle-Diffusion, which  enables fine-grained text-driven 3D stylization by guiding image rendering with a controllable 2D diffusion model in an end-to-end manner. We first briefly review the basic text-driven 3D stylization method (TANGO) and typical diffusion models. After that, we introduce the technical details of our proposed 3DStyle-Diffusion.
Figure \ref{fig:framework} demonstrates an overview of our approach.

\subsection{Preliminaries}\label{sec:preliminary}
{\bf Text-driven 3D Stylization}. Given an input mesh \bm{$M$} (consists of $e$ vertices $\bm{V} \in \mathcal{R}^{e\times3}$ and $u$ faces $\bm{F} \in \{1,...,n\}^{u\times3}$), one representative solution (TANGO \cite{chen2022tango}) is to parameterize its style as three learnable MLP networks, which present spatially varying BRDF (SVBRDF), normal and lighting properties respectively \cite{zhang2021physg}. Then stylized images can be generated with the learned implicit neural network parameters through a differentiable renderer. 
Specifically, to compute the color of each single pixel $p$ in the rendered image, a camera ray $\bm{R}_p=\bm{c}+t\bm{\nu}_p$ is emitted originating at the camera center $\bm{c}$ through the pixel $p$ along direction $\bm{\nu}_p$. The ray casting method \cite{roth1982ray} is then used to find the first intersection point $\bm{k}_p$ $\&$ intersection face $\bm{f}_p$ of the ray $\bm{R}_p$ and the mesh $\bm{M}$. Based on the intersections of camera ray and mesh, TANGO first uses a Normal network $\bm{\Pi}(\cdot)$ to estimate the normal vector $\hat{\bm{n}}_p$ of $\bm{k}_p$ according to the face normal $\bm{n}_p$ of $\bm{f}_p$ at point $\bm{k}_p$, i.e., $\hat{\bm{n}}_p=\bm{\Pi}(\bm{n}_p, \bm{k}_p)$. Next, TANGO utilizes a Lightning network $\bm{L}(\cdot)$ to predict the incident light density $\bm{L}_i(\bm{\omega}_i)$ from incident light direction $\bm{\omega}_i$. Lastly, a SVBRDF network $\bm{f}_r(\cdot)$ is leveraged to estimate the surface reflectance coefficients $\bm{f}_r(\bm{k}_p,\bm{\nu}_p,\bm{\omega}_i)$ of the material at location $\bm{k}_p$ from the viewing direction $\bm{\nu}_p$ and the incident light direction $\bm{\omega}_i$.
Accordingly, the final pixel color is represented as the observed light intensity $\bm{L}_p(\bm{\nu}_p, \bm{k}_p, \bm{n}_p)$, 
which is an integral over the hemisphere $\Omega = \left \{\bm{\omega}_i:\bm{\omega}_i\cdot\hat{\bm{n}}_p \geq 0 \right \}$:
\begin{equation}
  \bm{L}_p(\bm{\nu}_p, \bm{k}_p, \bm{n}_p) =
  \int_\Omega \bm{L}_i(\bm{\omega}_i)\bm{f}_r(\bm{k}_p,\bm{\nu}_p,\bm{\omega}_i)(\bm{\omega}_i \cdot 
  \hat{\bm{n}}_p ) \mathrm{d}\bm{\omega}_i.
\label{eq:rendering}
\end{equation}
To render an image $\bm{x} \in [0, 1]^{H \times W \times 3}$, a collection of rays are sampled corresponding to all the pixels in that image, and the resulting color values $\bm{L}_p$ are arranged into a 2D image. Note that the image rendering process is fully differentiable, which allows gradients to be backpropagated into the neural implicit representation network.

TANGO leverages a pre-trained CLIP model to optimize the Normal, Lightning, and SVBRDF networks jointly. CLIP is a powerful cross-modal model that comprises a text encoder $E_T$, and an image encoder $E_I$. Briefly, TANGO uses the following CLIP loss function as the overall objective:  
\begin{equation}
\label{eq:clip}
\mathcal{L}_{CLIP} = -E_{I}(\bm{x})^{T}E_{T}(y),
\end{equation}
where $y$ is the text prompt and $\bm{x}$ is the rendered image. Intuitively, CLIP loss holistically aligns the text prompt and rendered image in a shared latent embedding space at each training step.

{\bf Diffusion Models.} Diffusion models (DMs) are generative models that learn the data distribution from a Gaussian distribution through a gradual denoising process \cite{ho2020ddpm}. At each time step $t$, a \emph{forward diffusion 
process} $q(\mathbf{x}_t \vert \mathbf{x}_{t-1})$ is defined, which follows a Markov chain to gradually add a small amount of Gaussian noise to the sample $\mathbf{x}_0$ sampled from a real data distribution $\mathbf{x}_0 \sim q(\mathbf{x})$ (e.g., ``real images''). This produces a sequence of noisy samples after $T$ steps, $\mathbf{x}_1, \dots, \mathbf{x}_T$. The step sizes are controlled by a pre-determined variance schedule $0 < \beta_1 < \beta_2 < \dots < \beta_T < 1$:
\begin{equation}
  \begin{aligned}
    q(\mathbf{x}_t \vert \mathbf{x}_{t-1}) &= \mathcal{N}(\mathbf{x}_t; \sqrt{1 - \beta_t} \mathbf{x}_{t-1}, \beta_t\mathbf{I}).
  \end{aligned}
\end{equation}
After $T$ noise adding steps, $\mathbf{x}_T$ is equivalent to an isotropic Gaussian distribution. The {\em reverse diffusion process} aims to progressively ``denoise'' $\mathbf{x}_T$ to recover the true sample. 
To model the conditional probability for 
the reverse diffusion process, a neural network with parameters $\phi$ is used to approximate the distribution:
$p_\phi(\mathbf{x}_{t-1} \vert \mathbf{x}_t) = \mathcal{N}(\mathbf{x}_{t-1}; 
\boldsymbol{\mu}_\phi(\mathbf{x}_t, t), \boldsymbol{\Sigma}_\phi(\mathbf{x}_t, t))$.
As shown in \cite{ho2020ddpm}, noise approximation model $\boldsymbol{\epsilon}_\phi(\mathbf{x}_t, t)$ can be instead used to predict the noise contained in a noisy image $\mathbf{x}_t$ at time step $t$:
\begin{equation}
  \begin{aligned}
    \mathbf{x}_{t-1} = \mathcal{N}(\mathbf{x}_{t-1}; \frac{1}{\sqrt{\alpha_t}} 
    \Big( \mathbf{x}_t - \frac{1 - \alpha_t}{\sqrt{1 - \bar{\alpha}_t}} 
    \boldsymbol{\epsilon}_\phi(\mathbf{x}_t, t) \Big), \boldsymbol{\Sigma}_\phi(\mathbf{x}_t, t)),
  \end{aligned}
\end{equation}
where $\alpha_t = 1 - \beta_t$ and $\bar{\alpha}_t = \prod_{i=1}^t \alpha_i$ is the product of all $\alpha$ values up to time $t$.
For a conditional generation, such as in text-to-image diffusion models \cite{rombach2022stablediffusion}, 
a text prompt $y$ is embedded and used as a condition in the diffusion model via attention mechanism widely adopted in Vision Transformers \cite{yao2023dual,li2022contextual}. The corresponding noise predictor 
is denoted as $\boldsymbol{\epsilon}_\phi(\mathbf{x}_t;y,t)$ and the loss function is defined as follows:
\begin{equation}
  \mathcal{L}_\mathrm{diff}(\phi,x) = \mathbb{E}_{t,\epsilon}\Bigl[w(t)\|\boldsymbol{\epsilon}_\phi(\mathbf{x}_t, y,t) - \epsilon \|^2_2 \Bigr],
\label{equation:diffloss}
\end{equation}
where $w(t)$ is a weighting function depends on $t$ and $\epsilon \sim \mathcal{N}({\bf{0,I}})$.

\begin{figure}[tp]
\begin{center}
\includegraphics[width=0.98\linewidth]{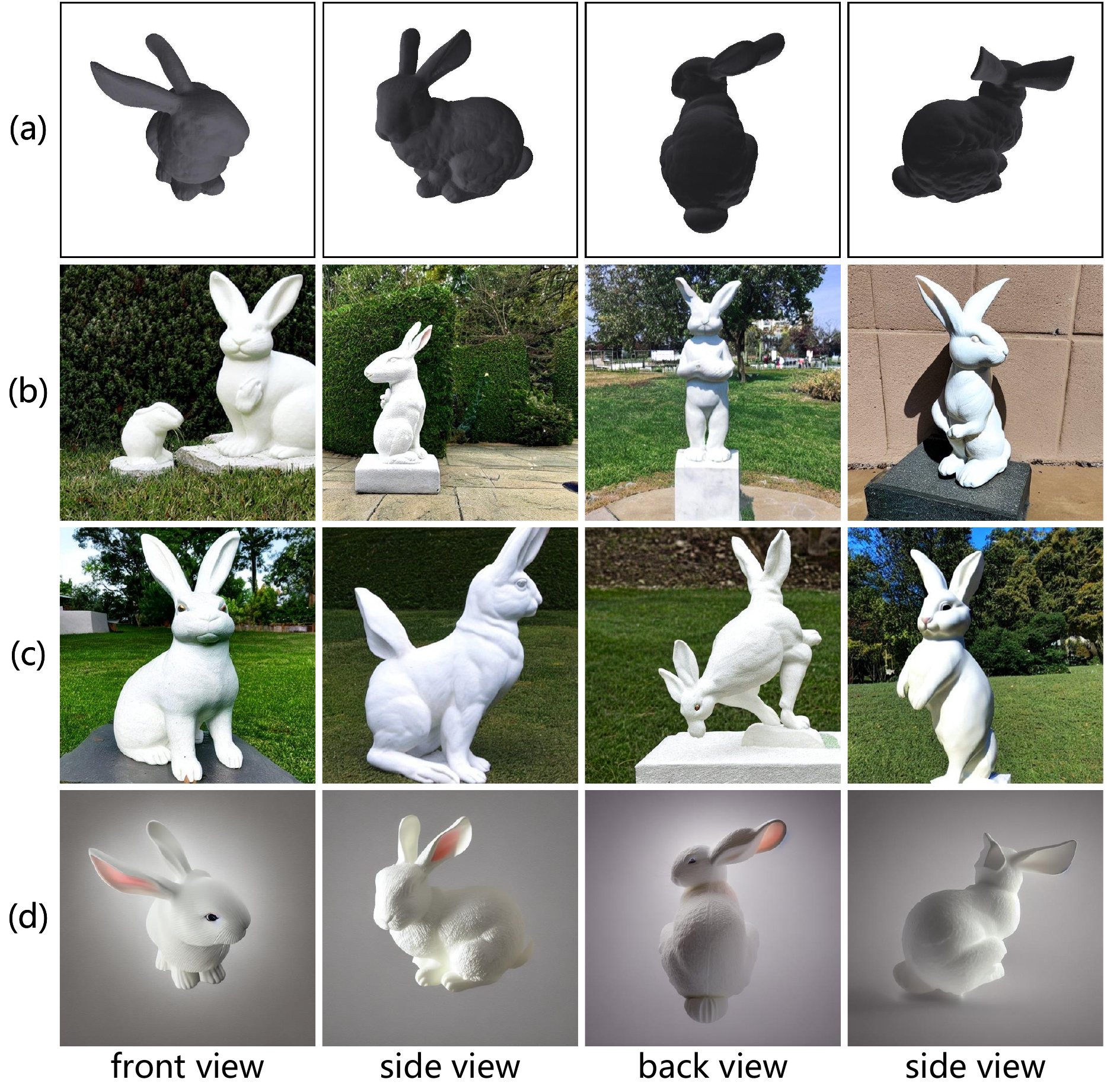}
\vspace{-0.05in}
\end{center}
\vspace{-0.1in}
  \caption{Motivation. (a) Rendered images of 3D mesh during training, (b) the results of Stable Diffusion under the same text prompt, (c) view-dependent prompting results of Stable Diffusion by augmenting text prompt with view names, (d) results of depth conditioned Stable Diffusion by introducing additional depth condition derived from 3D mesh. The input text prompt is ``a white statue of a rabbit''. Depth-conditioned results faithfully match the rendered images' pose and shape.}
\label{fig:motivation}
\vspace{-0.2in}
\end{figure}

\subsection{3DStyle-Diffusion}
{\bf Motivation.}
A recent pioneering practice (DreamFusion \cite{poole2022dreamfusion}) designs Score Distillation Sampling (SDS), which enables the use of a pre-trained text-to-image diffusion model to optimize a NeRF model \cite{mildenhall2020nerf, barron2022mipnerf360} solely based on
a text prompt. Specifically, at each training step, an image $x$ is first rendered by NeRF from a random camera viewpoint. DreamFusion then perturbs the rendered image $x$ with noise $\epsilon \sim \mathcal{N}({\bf{0,I}})$ into a noisy image $x_t = \sqrt{\bar{\alpha}}_{t}x+\sqrt{1-\bar{\alpha}}\epsilon$ according to the random sampled $t$. The noisy image is used to calculate the gradient to update NeRF parameters $\theta$: 
\begin{equation}
\label{eq:sds}
\nabla_{\theta}\mathcal{L}_{SDS}(\phi, x) = \mathbb{E}_{t,\epsilon}[ w(t)( \epsilon_\phi(x_t;t,y)-\epsilon)\frac{\partial x}{\partial \theta}], 
\end{equation}
where $y$ is the text prompt of the 3D scene to be generated. Intuitively, the SDS loss in Eq. \ref{eq:sds} pushes noisy versions of the rendered images to lower energy states of the text-to-image diffusion model. By randomly sampling views and backpropagating through the NeRF, it encourages the renderings to resemble images generated by the prior text-to-image diffusion model for a given text prompt.

By utilizing pre-trained text-to-image diffusion models, DreamFusion has demonstrated impressive text-to-3D generation results. Nevertheless, the CLIP model widely used in existing text-driven 3D stylization methods only holistically aligns images and texts in a semantic space, while text-to-image diffusion models can generate realistic images with fine-grained details based on text prompts.
Motivated by this, a natural question arises: \emph{is there an elegant way to trigger fine-grained text-driven 3D stylization with 2D diffusion models?} One simple and straight-forward way is to extend the score distillation sampling into 3D stylization of meshes by directly replacing the CLIP loss (Eq. \ref{eq:clip}) with SDS loss (Eq. \ref{eq:sds}). We name this run as TANGO+Fusion. However, we observe that such naive incorporation of SDS into TANGO results in unsatisfactory results (see Figure \ref{fig:ablation}). One key limitation might be the lack of 3D awareness for pre-trianed 2D diffusion models, which inevitably results in geometry inconsistent supervision for the rendered images across different views. As shown in Figure \ref{fig:motivation} (a), the rendered images of a 3D mesh are derived from arbitrary camera viewpoints during the training process and thus have various geometry shapes. When solely conditioned on the same text prompt, the 2D diffusion model (e.g., Stable Diffusion) always generates diverse images that don't match the viewpoint of the rendered images (Figure \ref{fig:motivation} (b)). To alleviate this issue, existing score distillation sampling-based works \cite{poole2022dreamfusion, wang2022score, metzer2022latent} all use a view-dependent prompting strategy. Specifically, they add ``back view'', ``side view'' or ``front view'' (according to the camera position) into the input text prompt to roughly describe the camera viewpoint. However, on the one hand, the input mesh in text-driven 3D stylization has an arbitrary pose. It is impossible to automatically identify which viewpoint is precisely corresponding to its ``front view'' unless we introduce additional cost to manually label it. On the other hand, this ad-hoc approach has severe limitations. As depicted in Figure \ref{fig:motivation} (c), the same text prompt (e.g., ``side view of a white statue of a rabbit'') can correspond to a wide range of different pose values. Due to the ambiguity in the interpretation of prompt from the same view, 3D stylization might result in distorted and unrealistic visuals that fail to match the original intent of the text prompt. Moreover, it is not trivial to generate images that faithfully match the specified view via the pre-trained diffusion models.

{\bf Depth-aware Score Distillation Sampling.} 
To mitigate the aforementioned issues, we propose a simple yet effective depth-aware score distillation sampling strategy for text-driven 3D stylization. Our launch point is to incorporate 3D awareness into score distillation sampling by leveraging the precise depth cues derived from primary 3D meshes.

Technically, in each training iteration, we render an image using the differentiable renderer from a randomly sampled camera 
position $\bm{c}$. Simultaneously, we can effortlessly obtain the corresponding depth map from the same camera position. As described in Sec \ref{sec:preliminary}, for each pixel $p$ in the rendered image, a camera ray $\bm{R}_p=\bm{c}+t\bm{\nu}_p$ is emitted that starts from $\bm{c}$ and points towards $p$. Then we find the first intersection point $\bm{k}_p$ between the ray $\bm{R}_p$ and the mesh $\bm{M}$ through ray casting \cite{roth1982ray}. The depth value of pixel $p$ can be calculated as follows:
\begin{equation}
\label{eq:depth}
  d_p = {\Vert \bm{k}_p - \bm{c} \Vert}_2 \cdot cos\gamma,
\end{equation}
where $\gamma$ is the angle between the ray $\bm{R}_p$ and another ray from camera $\bm{c}$ to the center pixel of the image. Thus we simultaneously render an image $\bm{x}$ and its corresponding depth map $\bm{d}$ according to Eq. \ref{eq:rendering} and Eq. \ref{eq:depth} respectively. We then remould the standard score distillation sampling by exploiting an image conditioned diffusion model (ControlNet \cite{zhang2023adding}) to trigger fine-grained text-driven 3D stylization. ControlNet is an end-to-end neural network architecture that controls large-scale pre-trained image diffusion models (Stable Diffusion) to learn task-specific input conditions. Specifically, herein we use ControlNet-depth\footnote{\url{https://huggingface.co/lllyasviel/sd-controlnet-depth}} as our diffusion prior model, which is trained on large-scale depth-image-text pairs and can enable depth-guided text-to-image generation. Since ControlNet-depth has two conditions (depth map $\bm{d}$ and text prompt $y$), the noise is estimated as follows:
\begin{equation}
\begin{aligned}
\hat{\epsilon}_\phi(x_t; t, y, \bm{d}) = &\epsilon_\phi(x_t; t, y, \bm{d}) \\ &+ s * (\epsilon_\phi(x_t; t, y, \bm{d}) - \epsilon_\phi(x_t; t)),
\end{aligned}
\end{equation}
where $s$ is a user-defined constant that controls the degree of guidance and $\epsilon_\phi(x_t; t)$ represents the noise prediction without conditioning \cite{ho2022classifier}. Similar to Eq. \ref{eq:sds}, the depth-aware score distillation sampling loss is defined as follows:
\begin{equation}
\label{eq:d-sds}
\nabla_{\theta}\mathcal{L}_{D-SDS}(\phi, x) = \mathbb{E}_{t,\epsilon}[ w(t)(\hat{\epsilon}_\phi(x_t;t,y, \bm{d})-\epsilon)\frac{\partial x}{\partial \theta}],
\end{equation}
where $\phi$ is the parameters of the pre-trained ControlNet-depth and $\theta$ is the parameters of learnable MLP networks. 

Intuitively, previous score distillation sampling (Eq. \ref{eq:sds}) roughly pushes rendered views of the stylized 3D mesh into a higher probability density region determined by the single text prompt, while neglecting the objects' geometrical poses and shapes. In contrast, with the help of depth guidance, our depth-aware score sampling distillation (Eq. \ref{eq:d-sds}) further narrows down the text-conditioned probability density into a more compact and precise region that also closely aligns the depth cues (see Figure \ref{fig:motivation} (d)). As such, our depth-aware score distillation sampling loss can provide more precise supervision during 3D stylization. In this way, we elegantly integrate the rendering of stylized 3D meshes and 2D diffusion process of controllable image synthesis into an end-to-end scheme, enabling a high-quality fine-grained stylization of 3D~meshes.

\textbf{Optimization.} Recall that the overall optimization of our 3DStyle-Diffusion is guided by a pre-trained 2D diffusion model \cite{zhang2023adding}. During each iteration, we sample one camera point and render an image $x$ $\&$ its corresponding depth map $\bm{d}$ by Eq. \ref{eq:rendering} and Eq. \ref{eq:depth} respectively. We then use the depth-aware score distillation sampling loss in Eq. \ref{eq:d-sds} as our overall objective.  

\begin{figure*}[htb]
\begin{center}
\includegraphics[width=0.98\linewidth]{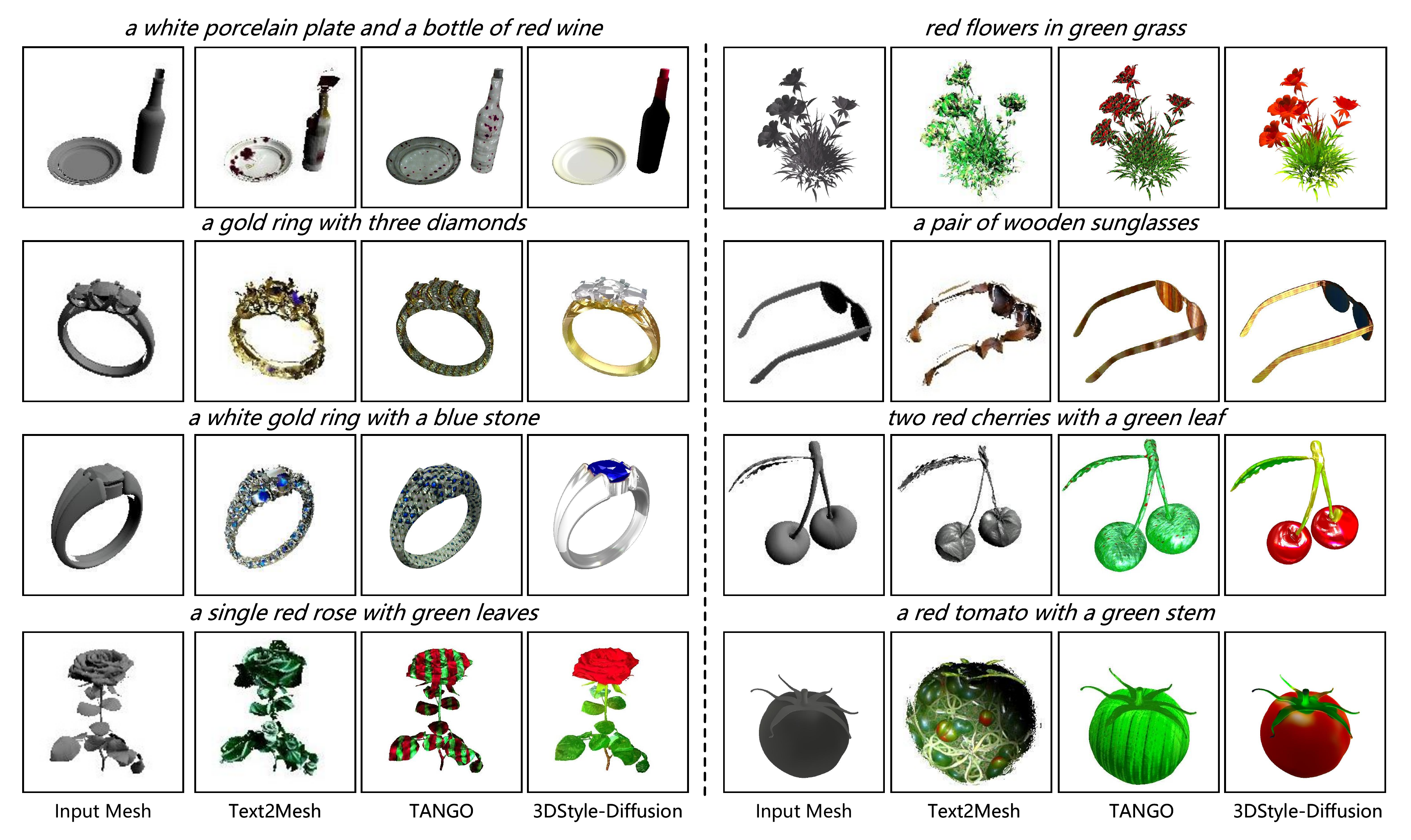}
\end{center}
\vspace{-0.18in}
\caption{Qualitative comparisons on text-driven 3D stylization. We compare our method with Text2Mesh \cite{Michel_2022_CVPR-Text2Mesh} and TANGO \cite{chen2022tango}. Our 3DStyle-Diffusion produces high fidelity and more realistic fine-grained stylization results.}
\label{fig:comparison_style}
\vspace{-0.15in}
\end{figure*}

\begin{figure*}[htb]
\begin{center}
\includegraphics[width=0.98\linewidth]{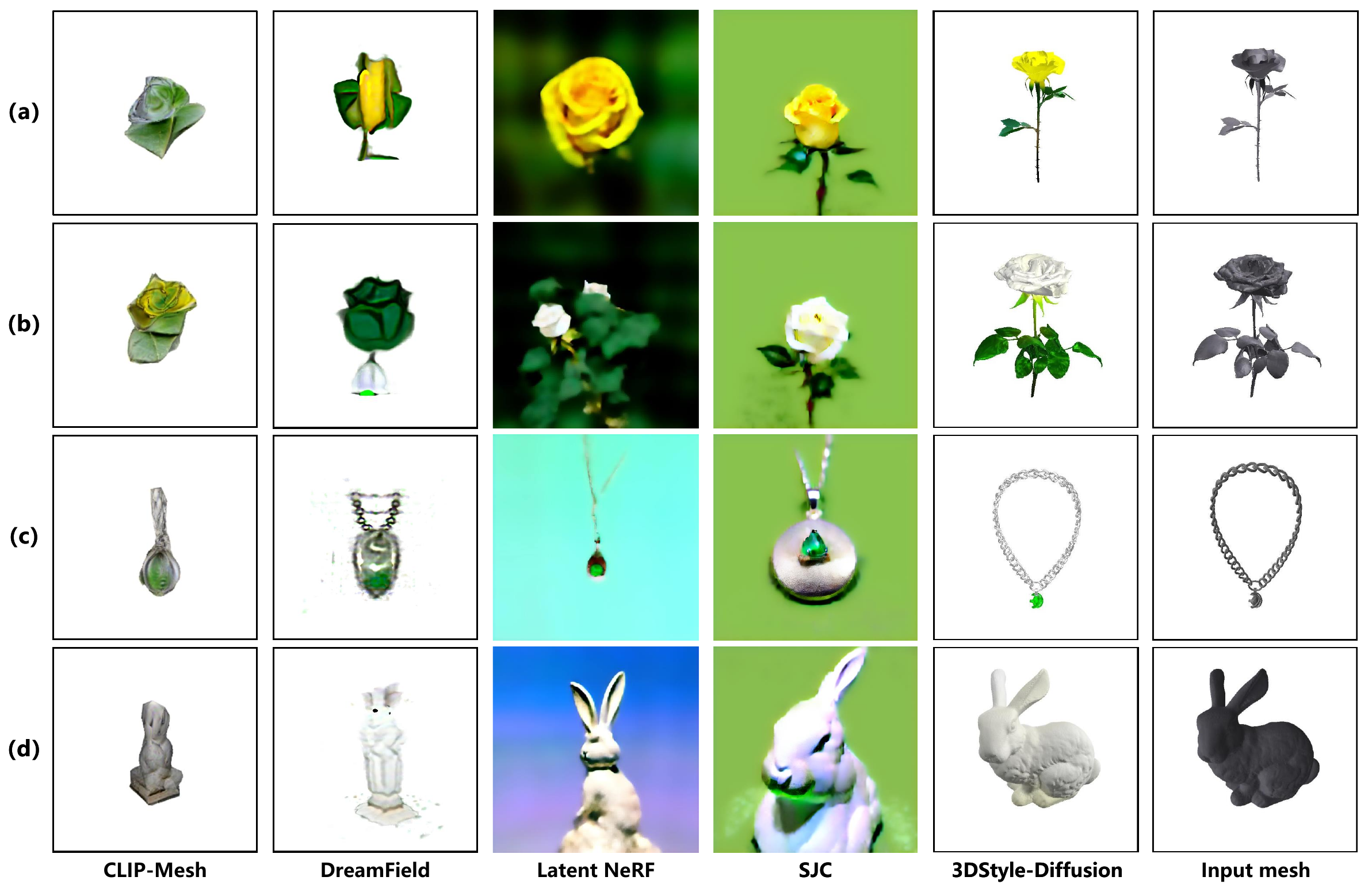}
\end{center}
\vspace{-0.1in}
\caption{Qualitative comparisons against zero-shot text-to-3D generation methods. We compare our method with CLIP-Mesh \cite{mohammad2022clip},  DreamField \cite{jain2022zero},  Latent-NeRF \cite{metzer2022latent} and Score Jacobian Chaining (SJC) \cite{wang2022score}.  The prompts are (a) ``a single yellow rose with green leaves''; (b) ``a single white rose with green leaves''; (c) ``a silver necklace with a green stone hanging from it''; (d) ``a white statue of a rabbit''. For each visualization of our 3DStyle-Diffusion, the corresponding input mesh is also visualized (the last column). Our 3DStyle-Diffusion shows better 3D results in terms of both geometry and texture in comparison to baselines.}
\label{fig:comparison_3d}
\vspace{-0.1in}
\end{figure*}

\section{Datasets and Metrics}
Existing text-driven 3D stylization works \cite{Michel_2022_CVPR-Text2Mesh, chen2022tango} commonly examine their methods over a limited number of meshes with simple text prompts. Meanwhile, they have only shown qualitative results of some cases and subjective user studies to evaluate the performances of the stylized 3D meshes. To alleviate these issues, we construct a new dataset that contains various challenging 3D composite assets (consisting of at least 2 different components) with more complex text prompts. A new evaluation protocol is also designed to quantitatively evaluate text-driven 3D stylization methods.

\textbf{Datasets.}
We build a new dataset of $107$ meshes derived from Objaverse \cite{deitke2022objaverse} (namely Objaverse-3DStyle), including various objects such as plants, animals, vehicles, etc. Although each mesh in Objaverse has a primary natural language description, we find these text descriptions are noisy and fail to accurately describe the mesh. To this end, we manually annotate each mesh with a precise text prompt. Specifically, given a 3D asset, we first uniformly render $36$ images from varied views. We then utilize a powerful vision-language model (BLIP \cite{li2022blip}) to automatically generate several captions according to the rendered images. Next, we manually check all the captions and rewrite the inaccurate captions. Finally, each mesh in our Objaverse-3DStyle has one ground truth text prompt plus $36$ ground truth rendered images. This dataset offers a fertile ground to evaluate text-driven 3D stylization methods.

\textbf{Evaluation Metrics.} \label{sec:metrics}
Following prior works \cite{poole2022dreamfusion, jain2022zero}, our evaluation metrics first involve the CLIP R-Precision, an metric that evaluates the consistency of rendered images with respect to the input text prompt. In general, CLIP R-Precision measures the precision by using CLIP to retrieve the correct caption from a set of distractors given a rendering of the stylized mesh. More specifically, we follow DreamField and employ a set of 153 prompts as distractors derived from the object-centric COCO \cite{lin2015microsoft} validation subset. We adopt CLIP ViT-B/32, CLIP ViT-B/16 and CLIP ViT-L/14 encoders to calculate R-Precsion. We also use an image-text CLIP score to measure the semantic similarity between rendered images and the input text prompt. Since we have ``ground truth'' stylized images in our built dataset, we further introduce two evaluation metrics: LPIPS \cite{zhang2018perceptual} and image-image CLIP score to assess the perceptual similarity and semantic similarity between rendered images and ``ground truth'' stylized images, respectively. Note that the encoder adopted in image-text and image-image CLIP scores is CLIP~ViT-L/14.

\section{EXPERIMENTS}
\textbf{Implementation Details.}
We implement the proposed 3DStyle-Diffusion mainly based on TANGO \cite{chen2022tango} codebase. Similarly, the Normal estimation network $\bm{\Pi}(\cdot)$ consists of 3 layers of width $256$. The SVBRDF network predicts diffuse, specular and roughness per surface point. We adopt the AdamW optimizer and the initial learning rate is set as $5 \times 10^{-4}$. The learning rate undergoes a decay of $0.7$ after every $500$ iterations. All experiments of 3DStyle-Diffusion are conducted on a single NVIDIA RTX 3090 GPU. We train the model for $3,000$ iterations and the whole training process takes approximately $0.5h$ for each mesh.

\textbf{Baselines.}
We compare against two state-of-the-art text-driven 3D stylization models (Text2Mesh \cite{Michel_2022_CVPR-Text2Mesh} and TANGO \cite{chen2022tango}). Text2Mesh is the pioneering practice, which leverages a CLIP model to predict stylized color and displacement for mesh vertexes. TANGO further incorporates reflectance properties and scene lighting to improve the realism of stylized meshes. We also compare against Geometry-Guided Latent-NeRF \cite{metzer2022latent}, which is mesh-guided text-to-3D generation. In addition, we include five typical zero-shot text-to-3D generation models as baselines. 1) CLIP-Mesh \cite{mohammad2022clip}, a method for generating 3D model from text prompt using a pre-trained CLIP \cite{radford2021clip} model. 2) DreamField \cite{jain2022zero}, which combines neural radiance fields with CLIP to synthesis diverse 3D objects form text prompt. 3) DreamFusion* \cite{threestudio2023}, a thrid-party implementation of DreamFusion \cite{poole2022dreamfusion}. 4) Latent-NeRF \cite{metzer2022latent}, which learns a NeRF model using a score distillation sampling loss in the latent space of Stable Diffusion. 5) SJC \cite{wang2022score}, is another SDS baesd text-to-3D framework.

\subsection{Qualitative Results}
The qualitative comparisons on text-driven 3D stylization are presented in Figure \ref{fig:comparison_style}. As shown in this figure, Text2Mesh is prone to produce stylized mesh with unreasonable deformation. For instance, given the text prompt \emph{``a red tomato with green stem''}, Text2Mesh generates severely distorted 3D shapes that do not conform to the original structure of the input mesh. TANGO doesn't suffer from the issue of geometry deformation, but fails to stylize the fine-grained components. Taking the third row (right half) in Figure \ref{fig:comparison_style} as an example, when using the text prompt \emph{“two red cherries with a green leaf”}, TANGO incorrectly generates completely green cherries. In contrast, we can clearly observe that the stylized 3D meshes of our 3DStyle-Diffusion faithfully respect both the fine-grained semantic context present in the input text prompt and the geometric structure specified in the input mesh.

\begin{table*}[htb]
\vspace{-0.0in}
  \caption{Quantitative comparisons on our constructed Objaverse-3DStyle benchmark. Our 3DStyle-Diffusion outperforms baseline models over most metrics. See section \ref{sec:metrics} for more details about the evaluation metrics adopted here.}
  \vspace{-0.1in}
  \begin{tabular}{l|ccc|cc|cc}
    \toprule
   \multirow{2}{*}{Method} &  \multicolumn{3}{c|}{CLIP R-Precision $\uparrow$} & \multicolumn{2}{c|}{CLIP Score $\uparrow$} & \multicolumn{2}{c}{LPIPS $\downarrow$}  \\
    &  ViT-B/32 &  ViT-B/16 &  ViT-L/14  &  Image-Text &  Image-Image & Alexnet &  VGG \\
    \midrule 
     SJC \cite{wang2022score} & 0.785 & 0.757  & 0.776  & 0.251  & 0.711 & 0.300 & 0.201 \\
      Latent-NeRF \cite{metzer2022latent} & 0.748  & 0.738  & 0.785  & 0.243  & 0.686 & 0.315 & 0.231 \\
       Dreamfusion* \cite{threestudio2023} & 0.776  & 0.757  & 0.794  & 0.254  & 0.732 & 0.309 & 0.221 \\
    Text2Mesh \cite{Michel_2022_CVPR-Text2Mesh} & \bf{0.841}  & 0.729  & 0.720  & 0.248  & 0.740 & 0.338 & 0.240 \\
    TANGO \cite{chen2022tango} & 0.804  &  0.794  &  0.813  & 0.245  &  0.768  & 0.295 & 0.204  \\
    Geometry-Guided Latent-NeRF \cite{metzer2022latent} & 0.822  & 0.804  & 0.832  & 0.248  & 0.819 & 0.270 & 0.188 \\
    TANGO+Fusion  & 0.813 &  0.804   & 0.822  & 0.250 & 0.817 & 0.272 & 0.191  \\
    3DStyle-Diffusion (ours) & 0.832  &  \bf{0.822}  & \bf{0.850}  &  \bf{0.260} & \bf{0.835}  & \bf{0.263} & \bf{0.182} \\
    \bottomrule
  \end{tabular}
  \vspace{-0.1in}
  \label{table:result}
\end{table*}

\begin{figure*}[htb]
\begin{center}
\includegraphics[width=1.0\linewidth]{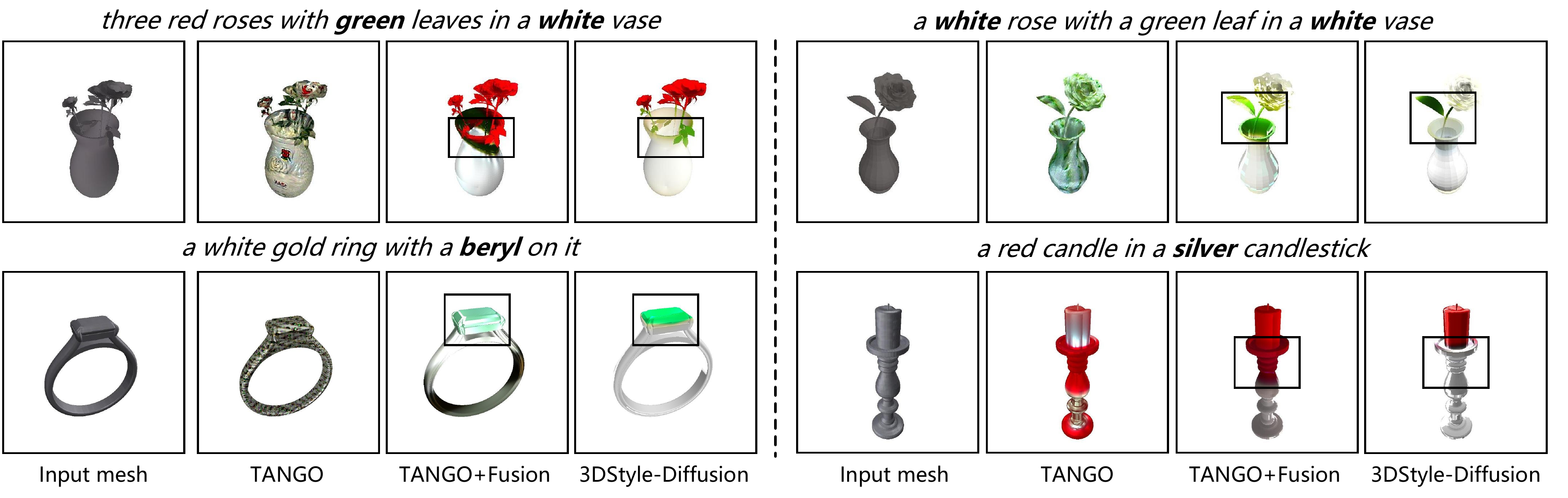}
\end{center}
\vspace{-0.1in}
\caption{Qualitative ablation study of our 3DStyle-Diffusion. TANGO \cite{chen2022tango} is our base model. TANGO+Fusion simply replaces the original CLIP loss in TANGO with score distillation sampling loss in DreamFusion \cite{poole2022dreamfusion}. Our 3DStyle-Diffusion clearly achieves better 3D stylization results than both ablated runs.}
\label{fig:ablation}
\vspace{-0.1in}
\end{figure*}

In addition, we depict the qualitative comparisons on zero-shot text-to-3D generation in Figure \ref{fig:comparison_3d}. As illustrated in this figure, existing zero-shot text-to-3D methods show inferior capability of 3D generation, making it difficult to generate precise and realistic 3D objects. Taking the first row (Figure \ref{fig:comparison_3d} (a)) as an example, when using the text prompt \emph{``a single yellow rose with green leaves''}, CLIP-Mesh and DreamField generate roses that are distorted and do not accurately match real-world roses' structures. Although Latent-NeRF and SJC can generate somewhat reasonable 3D shapes and visuals, they encounter challenges in the generation of fine-grained details, which consequently lead to unrealistic visual appearances. In contrast, our 3DStyle-Diffusion effectively generates higher-quality 3D results in terms of both geometry and texture by incorporating explicit 3D shape priors (i.e., the input mesh) and controllable 2D diffusion prior (i.e., depth-aware score distillation sampling).

\subsection{Quantitative Results}
The quantitative performance comparisons of different models for text-driven 3D stylization are summarized in Table \ref{table:result}. Overall, our 3DStyle-Diffusion consistently achieves better performances against state-of-the-art zero-shot text-to3D generation and text-driven 3D Stylization techniques over most metrics. The results generally highlight the key advantage of incorporating controllable 2D diffusion models into text-driven 3D stylization. It is worthy to note that both Text2Mesh and TANGO are directly optimized with CLIP ViT-B/32 during training, thereby leading to competitive CLIP R-Precision scores on ViT-B/32. However, our 3DStyle-Diffusion still manages to outperform all baselines over CLIP R-Precision scores under most backbones (ViT-B/16 $\&$ ViT-L/14) and image-text CLIP score. This demonstrates that our 3DStyle-Diffusion can generate high-quality content semantically aligned with the input text prompt. For image-image CLIP score and LPIPS, our 3DStyle-Diffusion again exhibits the best performances in comparison to all baselines, which demonstrates that our method can generate stylized meshes that faithfully match the target style.

\subsection{Ablation Study}
In this section, we investigate the effectiveness of our proposed depth-aware score distillation sampling in Eq. \ref{eq:d-sds}. We depict the qualitative results of each ablated run in Figure \ref{fig:ablation}. TANGO is the base model that leverages CLIP loss to perform text-driven 3D stylization. TANGO+Fusion is one degraded version of our 3DStyle-Diffusion by directly replacing the original CLIP loss in TANGO with score distillation sampling loss in DreamFusion \cite{poole2022dreamfusion}. As shown in this figure, TANGO+Fusion achieves better results than TANGO, which validates the effectiveness of score distillation sampling via typical diffusion models. Nevertheless, TANGO+Fusion still suffers from some unrealistic details of fine-grained components. In contrast, the results of 3DStyle-Diffusion precisely match the input text prompt and are faithfully photo-realistic. This clearly validates the merit of our designed depth-aware score distillation sampling. In addition, we also show the corresponding quantitative results of ablated runs in Table \ref{table:result}, which again validate the effectiveness of our proposal.

\subsection{User Study}
We additionally perform user study to evaluate our 3DStyle-Diffusion against two baseline models (i.e., Text2Mesh and TANGO) by comparing each pair. We invite 6 participants with different educational backgrounds and show them two videos (rotating 3D assets) side by side in each test case. The text prompt and corresponding input bare 3D mesh are also displayed to the participants. Each pair of videos is rendered from stylized meshes by two different methods using the same input bare mesh and text prompt. We then ask participants to choose the better one by jointly considering the following two aspects: (1) the alignment to the text prompt and (2) the fidelity of the visual appearance. According to all participants’ feedback, we measure the user preference score of 3DStyle-Diffusion as the percentage of its generated results that are preferred. Table \ref{table:user_study} shows the results of the user study. In general, our 3DStyle-Diffusion significantly outperforms both baselines with higher user preference rates. This validates the effectiveness of our proposed 3DStyle-Diffusion again in the aspect of human preference. 

\begin{table}[tb]
\caption{User Study. Users show a clear preference for our 3DStyle-Diffusion against Text2Mesh and TANGO.}
\centering
\vspace{-0.0in}
\begin{tabular}{lc}
\hline
Comparison & User Preference Score \\ \hline
3DStyle-Diffusion \emph{vs.} Text2Mesh &      85.0\%     \\
3DStyle-Diffusion \emph{vs.} TANGO  &      78.5\%     \\ \hline
\end{tabular}
\label{table:user_study}
\vspace{-0.15in}
\end{table}

\section{Conclusions}
In this paper, we have proposed 3DStyle-Diffusion, a novel method that enables more controllable stylization of 3D meshes with additional guidance from 2D Diffusion models. Specifically, our 3DStyle-Diffusion parameterizes the texture of 3D mesh into reflectance properties and scene lighting, and leverages a pre-trained controllable 2D diffusion model to guide the learning of rendered images, enabling a high-quality fine-grained stylization of 3D meshes. We also build a new challenging dataset derived from Objaverse and the evaluation protocol for this task. We validated our proposal through both qualitative and quantitative experiments and demonstrated its capability for 3D content creation via text-driven stylization. Furthermore, the ability in triggering fine-grained text-driven 3D stylization via 2D diffusion models is potentially a new paradigm of 3D content creation.

\begin{acks}
This project was supported by National Key R\&D Program of China (No. 2022YFB3104703) and in part by the National Natural Science Foundation of China (No. 62032006, 62172103).
\end{acks}


\end{document}